\documentclass{article}

\PassOptionsToPackage{numbers, compress}{natbib}

\usepackage[final]{neurips_2022}

\usepackage[utf8]{inputenc} 
\usepackage[T1]{fontenc}    
\usepackage{hyperref}       
\usepackage{url}            
\usepackage{booktabs}       
\usepackage{amsfonts}       
\usepackage{nicefrac}       
\usepackage{microtype}      
\usepackage{xcolor}         

\usepackage{graphicx}
\usepackage{svg}

\usepackage{amsmath}
\usepackage{dsfont}

\usepackage{multirow}
\usepackage{multicol}
\usepackage{subcaption}     
\usepackage{array}  
\usepackage{longtable}
\usepackage{comment}

\usepackage{listings}
\usepackage{algorithm}      
\usepackage[rightComments=false]{algpseudocodex}  

\usepackage{pdflscape}      
\usepackage[textsize=tiny]{todonotes}

\title{Rewards Encoding Environment Dynamics Improves Preference-based Reinforcement Learning}

\author{%
  Katherine Metcalf \quad Miguel Sarabia \quad Barry-John Theobald\\
 \{\texttt{kmetcalf}, \texttt{miguelsdc}, \texttt{barryjohn\_theobald}\}@apple.com\\
 Apple, California, USA.\\
}

\makeatletter
\def\@hangfrom#1{\setbox\@tempboxa\hbox{{#1}}%
      \hangindent 0pt
      \noindent\box\@tempboxa}
\makeatother

\begin{document}

\maketitle

\begin{abstract}
Preference-based reinforcement learning (RL) algorithms help avoid the pitfalls of hand-crafted reward functions by distilling them from human preference feedback, but they remain impractical due to the burdensome number of labels required from the human, even for relatively simple tasks.  In this work, we demonstrate that encoding environment dynamics in the reward function (REED) dramatically reduces the number of preference labels required in state-of-the-art preference-based RL frameworks. We hypothesize that REED-based methods better partition the state-action space and facilitate generalization to state-action pairs not included in the preference dataset. REED iterates between encoding environment dynamics in a state-action representation via a self-supervised temporal consistency task, and bootstrapping the preference-based reward function from the state-action representation. Whereas prior approaches train only on the preference-labelled trajectory pairs, REED exposes the state-action representation to all transitions experienced during policy training. We explore the benefits of REED within the PrefPPO \citep{christiano2017deep} and PEBBLE \citep{lee2021pebble} preference learning frameworks and demonstrate improvements across experimental conditions to both the speed of policy learning and the final policy performance. For example, on quadruped-walk and walker-walk with 50 preference labels, REED-based reward functions recover 83\% and 66\% of ground truth reward policy performance and without REED only 38\% and 21\% are recovered. For some domains, REED-based reward functions result in policies that outperform policies trained on the ground truth reward.
\end{abstract}

\section{Introduction}

The quality of a reinforcement learned (RL) policy depends directly on the quality of the reward function used to train it. However, hand-crafting reward functions is very challenging, even for experts.  A poorly specified reward can result in sub-optimal behaviors, with reward exploitation frequently leading to undesired behaviors \cite{bostrom2014superintelligence,amodei2016concrete,hadfield2017inverse}. Therefore, methods for specifying robust, non-hackable reward functions are critical to correctly and efficiently train policies with RL. 
One such method is to teach an agent to align its policy with human preferences by distilling reward functions from feedback on sets of trajectories \citep{christiano2017deep,lee2021pebble,ibarz2018reward, hadfield2016cooperative,leike2018scalable,stiennon2020learning,wu2021recursively,lee2021bpref,park2022surf}.

Learning from preference labels over sets of trajectories is promising as it empowers non-experts to provide such feedback, especially compared to providing corrections (e.g.\ DAgger \citep{ross2011reduction} and related approaches) or providing real-valued rewards (e.g.\ TAMER \citep{knox2008tamer} and related approaches). Still, the majority of existing preference-based deep RL methods require either a dataset of demonstrations \cite{ibarz2018reward} or thousands of preference labels to recover optimal policy performance \cite{christiano2017deep,lee2021pebble,park2022surf}.

In this paper, we target efficient reward function learning by introducing \textbf{R}ewards \textbf{E}ncoding \textbf{E}nvironment \textbf{D}ynamics (REED) (Section \ref{sec:encoding_env_dynamics}). Given the difficulty people face when providing feedback for a single state-action pair (e.g.\ \cite{knox2008tamer}), and the importance of defining preferences over transitions instead of single state-action pairs \citep{lee2021pebble}, \emph{it is likely that people's internal reward functions are defined over outcomes not state-action pairs}.

We hypothesize that: (1) modelling the relationship between state, action, and next-state triplets is essential for learning preferences over transitions, (2) encoding environment dynamics with a temporal consistency objective will allow the reward function to better generalize over states and actions with similar outcomes, and (3) exposing the reward model to all transitions experienced by the policy during training will result in more stable reward estimations during reward and policy learning. Therefore, we incorporate a self-supervised temporal consistency task using self-predictive representations (SPR) \cite{schwarzer2020data} into preference-based RL frameworks. REED reward functions lead to faster policy training and reduce the number of preference samples needed (as we will show in Section \ref{sec:experiments}).  

We demonstrate the benefits of REED using the current state-of-the-art in preference learning (\citep{christiano2017deep}, \citep{lee2021pebble}). In our experiments, which follow those outlined in \citep{lee2021bpref}, REED reward functions outperform non-REED reward functions across different preference dataset sizes, quality of preference labelling strategy, and tasks (Section \ref{sec:joint_results}). The improvements in policy learning and preference sample efficiency represent a significant step towards enabling end-users to naturally adjust the behavior of agents in their environment with their own feedback.

\section{Related Work}

\textbf{Learning from Human Feedback.} Learning reward functions from preference-based feedback \cite{akrour2011preference,akrour2012april,wilson2012bayesian,sugiyama2012preference,wirth2013preference,wirth2016model,sadigh2017active,leike2018scalable} has been used to address the limitations of learning policies directly from human feedback \cite{pilarski2011online,macglashan2017interactive,arumugam2019deep} by inferring reward functions from either task success \cite{zhang2019solar,singh2019end,smith2019avid} or real-valued reward labels \cite{knox2009interactively,warnell2018deep}. Learning policies directly from human feedback is time inefficient for the human as near constant supervision is frequently assumed.  Inferring reward functions from task success feedback requires examples of task success, which can be difficult to acquire in complex and multi-step task domains.  Finally, people are not able to reliably provide real-valued reward labels. Preference-based RL was extended to deep RL domains in \citep{christiano2017deep} then made more efficient and improved in \citep{lee2021pebble} and \citep{park2022surf}. To reduce the feedback complexity of preference-based RL, \citet{lee2021pebble} sped up policy learning via (1) intrinsically-motivated exploration and (2) relabelling the experience replay buffer. These two techniques aimed to improve both the sample complexity of the policy, and the trajectories generated by the policy, which are then used to seek feedback. \citet{park2022surf} reduced feedback complexity by incorporating augmentations and pseudo-labelling into the reward model learning. Additionally, preference-learning has also been incorporated into data-driven skill extraction and execution in the absence of a known reward function \cite{wang2021skill}. 

\textbf{Encoding Environment Dynamics.} Prior work has demonstrated the benefits of encoding environment dynamics in the state-action representation of a policy \cite{schwarzer2020data,zhang2020learning,ye2021mastering}. We instead encode environment dynamics in the reward function which, as far as we are aware, has not been done before. It is nonetheless common for dynamics models to predict both the next state and the environment's reward \cite{zhang2020learning} which suggests it is important to imbue the reward function with some understanding of the dynamics. The primary self-supervised approach to learning a dynamics model is to predict the latent next state \cite{schwarzer2020data,zhang2020learning,hafner2019learning,lee2020stochastic}. Indeed, the current state-of-the-art in data efficient RL \cite{schwarzer2020data,ye2021mastering} for encoding dynamics via latent next-state predictions uses a self-predictive representation (SPR) \cite{schwarzer2020data} task, an approach we follow in this work.

\section{Preference-based Reinforcement Learning} \label{sec:pref_learning}

Reinforcement learning (RL) trains an agent to achieve tasks via environment interactions and reward signals \cite{sutton2018reinforcement}. For each time step \(t\) the environment provides a state \(s_{t}\) used by the agent to select an action according to its policy \(a_{t} \sim \pi_{\phi}(a|s_{t})\). Then \(a_{t}\) is applied to the environment, which returns: a next state according to its transition function \(s_{t+1} \sim \tau(s_{t}, a_{t})\), and a reward \(r(s_{t}, a_{t})\). The agent's goal is to learn a policy \(\pi_{\phi}\) maximizing the expected discounted return, \(\sum_{k=0}^{\infty}\gamma^{k}r(s_{t+k}, a_{t+k})\).
In preference-based RL \cite{christiano2017deep,ibarz2018reward,leike2018scalable,stiennon2020learning,wu2021recursively,lee2021pebble,lee2021bpref} \(\pi_{\phi}\) is trained with a reward function \(\hat{r}_{\psi}\) distilled from preferences \(P_{\psi}\) iteratively queried from a teacher, where \(r_{\psi}\) is assumed to be a latent factor explaining the preference \(P_{\psi}\). A buffer \(\mathcal{B}\) of transitions is accumulated as \(\pi_{\phi}\) learns and explores.

A labelled preference dataset \(\mathcal{D}_\text{pref}\) is acquired by querying a teacher for preference labels every \(K\) steps of policy training and is stored as triplets \((\sigma^{1}, \sigma^{2}, y_{p})\), where \(\sigma^{1}\) and \(\sigma^{2}\) are trajectory segments (sequences of state-action pairs) of length \(l\), and \(y_{p}\) is a preference label indicating which, if any, of the trajectories is preferred \cite{lee2021bpref}. To query the teacher, the $M$ \textit{maximally informative} pairs of trajectory segments (e.g.\ pairs that most reduce model uncertainty as measured by reward ensemble disagreement) are sampled from \(\mathcal{B}\), sent to the teacher for preference labelling, and stored in \(\mathcal{D}_\text{pref}\) \cite{lee2021bpref,sadigh2017active,biyik2018batch,biyik2020active}. Typically \(\mathcal{D}_\text{pref}\) is used to update \(\hat{r}_{\psi}\) on a schedule conditioned on the training steps for \(\pi_{\phi}\)  (e.g.\ every time the teacher is queried).

The preference triplets \((\sigma^{1}, \sigma^{2}, y_{p})\) create a supervised preference prediction task to approximate \(r_{\psi}\) with \(\hat{r}_{\psi}\)  \citep{wilson2012bayesian,christiano2017deep,lee2021pebble}. The prediction task follows the Bradley-Terry model for a stochastic teacher \cite{bradley1952rank} and assumes that the preferred trajectory has a higher cumulative reward according to the teacher's \(r_{\psi}\). The probability of the teacher preferring \(\sigma^{1}\) over \(\sigma^{2}\) (\(\sigma^{1} \succ \sigma^{2}\)) is formalized as:
\begin{equation}
    P_{\psi}[\sigma^{1} \succ \sigma^{2}] = \frac{\exp\sum_{t}{\hat{r}_{\psi}(s^{1}_{t}, a^{1}_{t})}}{\sum_{i \in \{1, 2\}}{\exp \sum_{t}{\hat{r}_{\psi}(s^{i}_{t}, a^{i}_{t})}}},
\label{eq:preference_probability}
\end{equation}
where $s^{i}_{t}$ is the state at time step $t$ of trajectory $i \in \{1, 2\}$, and $a^{i}_{t}$ is the corresponding action taken.  

The parameters \(\psi\) of \(\hat{r}_{\psi}\) are optimized such that the binary cross-entropy over \(\mathcal{D}_\text{pref}\) is minimized:
\begin{equation}
    \mathcal{L}^{\psi} = \mathop{\mathbb{-E}}_{(\sigma^{1}, \sigma^{2}, y_{p}) \sim \mathcal{D}_\text{pref}}y_{p}(0) \log P_{\psi}[\sigma^{2} \succ \sigma^{1}] + y_{p}(1) \log P_{\psi}[\sigma^{1} \succ \sigma^{2}].
\label{eq:pref_objective}    
\end{equation}

While \(P_{\psi}[\sigma^{1} \succ \sigma^{2}]\) and \(\mathcal{L}^{\psi}\) are defined over trajectory segments, \(\hat{r}_{\psi}\) operates over individual \((s_{t}, a_{t})\) pairs. Each reward estimation in Equation \ref{eq:preference_probability} is made \emph{independently} of the other \((s_{t}, a_{t})\) pairs in the trajectory and \(P_{\psi}[\sigma^{1} \succ \sigma^{2}]\) simply sums the independently estimated rewards. Therefore, environment dynamics, or the outcome of different actions in different states, are not explicitly encoded in the reward function limiting its ability to model the relationship between state-action pairs and the values associated with their outcomes. By supplementing the supervised preference prediction task with a self-supervised temporal consistency task (Section \ref{sec:encoding_env_dynamics}), we can take advantage of all transitions experienced by \(\pi_{\phi}\) to learn a state-action representation in a way that: explicitly encodes environment dynamics, and can be used to learn to solve the preference prediction task.

\section{Encoding Environment Dynamics in the Reward Function}\label{sec:encoding_dynamics}

In this section, we present our approach to encoding environment dynamics via a temporal consistency task into the state-action representation of a preference-learned reward function. Our approach can be used with any existing preference-based RL approach. The main idea is to apply a self-supervised temporal consistency task to all transitions stored in \(\mathcal{B}\) to learn a state-action representation that is predictive of the latent representation of the next state. Preferences are then learned with a linear layer over the state-action representation.

\subsection{Rewards Encoding Environment Dynamics (REED)} \label{sec:encoding_env_dynamics}

\begin{figure}[ht]
    \centering
    \includegraphics[width=0.99\textwidth]{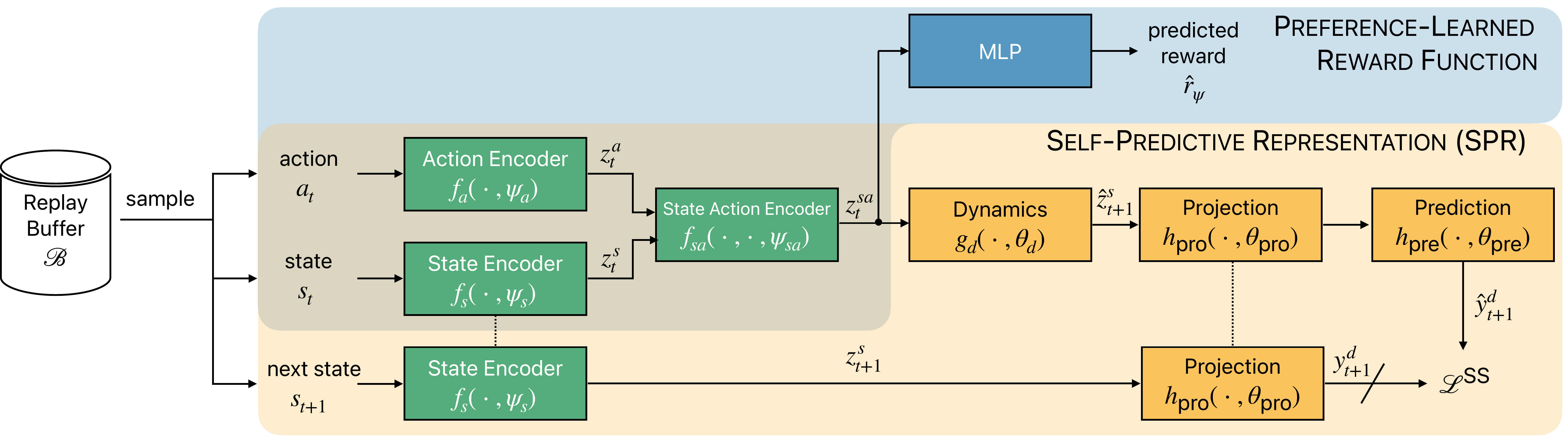}
    \caption{Architecture diagrams for self-predictive representation (SPR) objective \cite{schwarzer2020data} (in yellow), and preference-learned reward function (in blue). Modules in green are shared between SPR and the preference-learned reward function.}
    \label{fig:network_architecture}
\end{figure}

We use a self-supervised temporal consistency task based on SPR \cite{schwarzer2020data} to learn state-action representations that are predictive of likely future states and thus environment dynamics. The state-action representations are then bootstrapped to solve the preference prediction task in Equation \ref{eq:pref_objective} (see Figure \ref{fig:network_architecture} for an overview of the architecture). The SPR network is parameterized by \(\psi\) and \(\theta\), where \(\psi\) is shared with \(\hat{r}_{\psi}\) and $\theta$ is unique to the SPR network. At train time, batches of ($s_{t}$, $a_{t}$, $s_{t+1}$) triplets are sampled from a buffer $\mathcal{B}$ and encoded: \(f_{s}(s_{t}, \psi_{s}) \rightarrow z^{s}_{t}\), \(f_{a}(a_{t}, \psi_{a}) \rightarrow z^{a}_{t}\), \(f_{sa}(z^{s}_{t}, z^{a}_{t}, \psi_{sa}) \rightarrow z^{sa}_{t}\), and \(f_{s}(s_{t+t}, \psi_{s}) \rightarrow z^{s}_{t+1}\).  The embedding \(z^{s}_{t+1}\) is used to form our target for Equations \ref{eq:simsiam_objective} and \ref{eq:contrastive_objective}. A dynamics function \(g_{d}(z^{sa}_{t}, \theta_{d}) \rightarrow \hat{z}^{s}_{t+1}\) then predicts the latent representation of the next state \(z^{s}_{t+1}\). The functions \(f_{s}(\cdot)\), \(f_{a}(\cdot)\), and \(g_{d}(\cdot)\) are multi-layer perceptrons (MLPs), and \(f_{sa}(\cdot)\) concatenates \(z^{s}_{t}\) and \(z^{a}_{t}\) along the feature dimension before encoding them with a MLP. To encourage knowledge of environment dynamics in \(z^{sa}_{t}\), \(g_{d}(\cdot)\) is kept simple, e.g.\ a linear layer.

Following \cite{schwarzer2020data}, a projection head \(h_{\text{pro}}(\cdot, \theta_{\text{pro}})\) is used to project both the predicted and target next state representations to smaller latent spaces via a bottleneck layer and a prediction head \(h_{\text{pre}}(\cdot, \theta_{\text{pre}})\) is used to predict the target projections: \(\hat{y}^{d}_{t+1} = h_{\text{pre}}(h_{\text{pro}}(\hat{z}_{t+1}, \theta_{\text{pro}}), \theta_{\text{pre}})\) and \(y^{d}_{t+1} = h_{\text{pro}}(z_{t+1}, \theta_{\text{pro}})\). Both \(h_{\text{pro}}\) and \(h_{\text{pre}}\) are modelled as linear layers.

The benefits of REED should be independent of self-supervised objective function. Therefore, we present results for both a SimSiam (\(\mathcal{L}^{\text{SS}}\)) \cite{chen2021exploring,ye2021mastering} and a contrastive (\(\mathcal{L}^{\text{C}}\)) \cite{chen2020simple,oord2018representation,mazoure2020deep} loss. In the SimSiam loss, a stop gradient operation, \(\text{sg}(\text{...})\), is applied to $y^{d}_{t+1}$ and then $\hat{y}^{d}_{t+1}$ is pushed to be consistent with $z^{s}_{t+1}$ via a negative cosine similarity loss. In the contrastive loss, a stop gradient operation is applied to $y^{d}_{t+1}$ and then $\hat{y}^{d}_{t+1}$ is pushed to be predictive of which candidate next state is the true next state via the NT-Xent loss.
\noindent\begin{minipage}{.35\linewidth}
\begin{equation}
    \mathcal{L}^{\text{SS}} = - \cos(\hat{y}^{d}_{t+1}, \text{sg}(y^{d}_{t+1}))
\label{eq:simsiam_objective}
\end{equation}
\end{minipage}%
\begin{minipage}{.65\linewidth}
\begin{equation} 
    \mathcal{L}^{\text{C}} = -\log \frac{\exp(\cos(\hat{y}^{d}_{t+1}, \text{sg}(y^{d}_{t+1})) / \tau)}{\sum_{k=1}^{2N}\mathds{1}_{[s_{k} \neq s_{t+1}]} \exp(\cos(y^{d}_{t+1}, \hat{y}^{d}_{k}) \tau)}
\label{eq:contrastive_objective}
\end{equation}
\end{minipage}

Rather than applying augmentations to the input, as is typical in SimSiam and contrastive learning, temporally adjacent states are used to create the different views \cite{schwarzer2020data,ye2021mastering,oord2018representation,mazoure2020deep}. The architectures for each function composing the SPR network are specified in Section \ref{app_sec:architecture_details}.

\textbf{State-Action Fusion Reward Network.} REED requires a modification to the reward network architecture used by \citep{christiano2017deep} and \citep{lee2021pebble} as latent \emph{state} representations are compared and the next state representation (\(z^{s}_{t+1}\)) must be independent of next action. If \(z^{s}_{t+1}\) is not independent of the next action then the reward function will \emph{not} be independent of the policy. Instead of concatenating the raw state-action features, we separately encode the state, \(f_{s}(\cdot)\), and action, \(f_{a}(\cdot)\), before concatenating the embeddings and passing them to the body of our reward network. Separately encoding the state and action features adds extra parameters to the network, but the increase is small relative to overall network size (e.g., $\approx0.007\%$ additional parameters for the walker-walk task). For the purposes of comparison, we refer to the modified reward network as the state-action fusion (SAF) reward network. For architecture details, see Section \ref{app_sec:architecture_details}.

\subsection{Extending Preference Learning with REED}

The REED self-supervised temporal consistency task is used to update the parameters \(\psi\) and \(\theta\) each time the reward network is updated (every \(K\) steps of policy training, Section~\ref{sec:pref_learning}). All transitions in the buffer \(\mathcal{B}\) are used to update the state-action representation \(z^{sa}\), which effectively increases the amount of data used to train the reward function from \(M \cdot K\) preference triplets to all state-action pairs experienced by the policy \footnote{Note the reward function is still trained with \(M \cdot K\) triplets, but the state-action encoder has the opportunity to better capture the dynamics of the environment.}. REED precedes selecting and presenting the $M$ queries to the teacher for feedback. Updating \(\psi\) and \(\theta\) \emph{prior} to querying the teacher exposes \(z^{sa}\) to a larger space of environment dynamics (all transitions collected since the last model update), which enables the model to learn more about the world prior to selecting informative trajectory pairs for the teacher to label. The state-action representation \(z^{sa}\) plus a linear prediction layer is used to solve the preference prediction task (Equation \ref{eq:pref_objective}). After each update to \(\hat{r}_{\psi}\), \(\pi_{\phi}\) is trained on the updated \(\hat{r}_{\psi}\). See Appendix \ref{app_sec:pebble_pretrain_sfc_algorithm} for REED incorporated into PrefPPO \cite{christiano2017deep} and PEBBLE \cite{lee2021pebble}.

\section{Experiments} \label{sec:experiments}

Our experimental results in Section \ref{sec:joint_results} demonstrate that explicitly encoding environment dynamics in the reward function improves policy performance within preference-based RL.

\begin{table*}[t]
\caption{Ratio of policy performance (Equation \ref{eq:normalized_returns}) on learned versus ground truth rewards with the oracle labelling style with disagreement sampling. The results are reported as averages over 10 random seeds and the best results per preference-learning framework is highlighted in \textbf{bold}. \textsc{Base} refers to the baseline PEBBLE and PrefPPO methods, whereas \textsc{+SimSiam} is the SimSiam version of REED, and \textsc{+Contrast.} is the contrastive version of REED.}
\label{tab:perf_ratio_summary}
\begin{center}
\begin{footnotesize}
\begin{sc}
\hspace*{-1.3cm}
\begin{tabular}{m{1.5cm} @{\hspace{0.75cm}} l c c c @{\hspace{0.75cm}} c c c}
\toprule
\multirow{2}{*}{Task} & \multirow{2}{*}{Feed.} & \multicolumn{3}{c}{PEBBLE} & \multicolumn{3}{c}{PrefPPO}\\
 &  & Base & +SimSiam & +Contrast. & Base & +SimSiam & +Contrast. \\
\midrule
\multirow{3}{*}{\parbox{1.5cm}{Walker walk}} & 500 & \(0.74\pm0.18\) & \(0.86\pm0.20\) & \(\mathbf{0.9\pm0.17}\) & \(\mathbf{0.95\pm0.05}\) & $0.88\pm0.07$ & $0.93\pm0.06$ \\
    & 100 & $0.34\pm0.11$ & $0.68\pm0.23$ & \(\mathbf{0.78\pm0.21}\) & $0.68\pm0.08$ & $0.67\pm0.08$ & \(\mathbf{0.72\pm0.08}\) \\
    & 50 & $0.21\pm0.10$ & \(\mathbf{0.66\pm0.24}\) & $0.62\pm0.22$ & $0.51\pm0.13$ & \(\mathbf{0.58\pm0.13}\) & \(\mathbf{0.58\pm0.12}\) \\
\midrule
\multirow{3}{*}{\parbox{1.5cm}{Quadruped walk}}  & 500 & $0.56\pm0.21$ & \(\mathbf{1.1\pm0.21}\) & \(\mathbf{1.1\pm0.21}\) & $0.8\pm0.18$ & $1.1\pm0.20$ & $1.1\pm0.21$  \\
  & 100 & $0.38\pm0.21$ & \(\mathbf{0.78\pm0.16}\) & $0.67\pm0.18$ & $0.56\pm0.31$ & $1.0\pm0.24$ & $0.91\pm0.19$ \\
  & 50 & $0.38\pm0.26$ & $0.65\pm0.16$ & \(\mathbf{0.83\pm0.12}\) & $0.68\pm0.30$ & $0.9\pm0.19$ & \(1.2\pm0.34\) \\
\midrule
\multirow{3}{*}{\parbox{1.5cm}{Cheetah run}}  & 500 & \(0.86\pm0.14\) & \(0.88\pm0.22\) & \(\mathbf{0.94\pm0.21}\) & \(0.62\pm0.04\) & \(\mathbf{0.67\pm0.06}\) & \(0.66\pm0.06\) \\
  & 100 & \(0.4\pm0.14\) & \(\mathbf{0.69\pm0.26}\) & \(0.64\pm0.28)\) & \(0.46\pm0.04\) & \(0.49\pm0.04\) & \(\mathbf{0.54\pm0.04}\)\\
  & 50 & \(0.35\pm0.11\) & \(0.63\pm0.23\) & \(\mathbf{0.7\pm0.28}\) & \(\mathbf{0.5\pm0.07}\) & \(0.44\pm0.04\) & \(0.47\pm0.05\)\\
\midrule
\multirow{3}{*}{\parbox{1.5cm}{Button Press}} & 10k & \(\mathbf{0.66\pm0.26}\) & \textit{Collapses} & \(0.65\pm0.27\) & \(\mathbf{0.18\pm0.03}\) & \textit{Collapses} & \(0.15\pm0.04\) \\
  & 5k & \(0.48\pm0.21\) & \textit{Collapses} & \(\mathbf{0.55\pm0.24}\) & \(\mathbf{0.15\pm0.04}\) & \textit{Collapses} & \(0.14\pm0.04\) \\
  & 2.5k & \(0.37\pm0.18\) & \textit{Collapses} & \(\mathbf{0.49\pm0.25}\) & \(\mathbf{0.14\pm0.04}\) & \textit{Collapses} & \(\mathbf{0.14\pm0.04}\) \\
 \midrule
\multirow{3}{*}{\parbox{1.5cm}{Sweep Into}}  & 10k & \(0.28\pm0.12\) & \textit{Collapses} & \(\mathbf{0.47\pm0.23}\) & \(\mathbf{0.16\pm0.05}\) & \textit{Collapses} & \(0.11\pm0.03\) \\
  & 5k & \(0.17\pm0.01\) & \textit{Collapses} & \(\mathbf{0.34\pm0.14}\) & \(0.1\pm0.04\) & \textit{Collapses} & \(\mathbf{0.14\pm0.05}\) \\
  & 2.5k & \(0.15\pm0.09\) & \textit{Collapses} & \(\mathbf{0.21\pm0.13}\) & \(\mathbf{0.092\pm0.03}\) & \textit{Collapses} & \(0.058\pm0.02\) \\
\bottomrule
\end{tabular}
\end{sc}
\end{footnotesize}
\end{center}
\end{table*}

\subsection{Experimental Setup}\label{sec:experimental_setup}

We evaluate on the DeepMind Control Suite (DMC) \cite{tassa2018deepmind} and Meta-World \cite{yu2020meta} environments. DMC provides locomotion tasks with varying degrees of difficulty, while MetaWorld contains object manipulation tasks. For each task from DMC and MetaWorld, we evaluate performance on varying amounts of teacher feedback, i.e.\ different preference dataset sizes, and different teacher labelling strategies. The number of queries ($M$) presented to the teacher every $K$ steps is set such that for a given task, teacher feedback always stops at the same episode across the different total teacher feedback amounts. Feedback is provided by simulated teachers following \cite{christiano2017deep,lee2021pebble, lee2021bpref,park2022surf,wang2021skill}, where six different labelling strategies are used to evaluate model performance in the face of different types and amounts of labelling noise. These teaching strategies were first proposed as a configurable stochastic preference model in \cite{lee2021bpref}. An overview of the labelling strategies is provided in Appendix \ref{app_sec:labelling_strategies}.

Following \cite{christiano2017deep} and \cite{lee2021pebble}, an ensemble of three networks is used to model \(\hat{r}_{\psi}\) with a corresponding ensemble for the REED temporal consistency task. The ensemble is key for disagreement-based query sampling (see Section \ref{app_sec:disagreement_sampling}) and was shown in \cite{lee2021bpref} to improve final policy. All queried segments are of a fixed length (\(l=50\))\footnote{Fixed segments lengths are not strictly necessary. However, when evaluating with simulated humans, the importance of fixed length segments depends on the environment ground truth reward function. For example, variable length segments are necessary when the reward is a constant step penalty.}. The Adam optimizer \cite{kingma2015adam} with (\(\beta_{1}=0.9\), \(\beta_{2}=0.999\), \& no $L_{2}$-regularization) \cite{pytorch2019} is used to train on both the REED and non-REED tasks. For all PEBBLE-related methods, intrinsic policy training is reported in the learning curves and occurs over the first 9000 steps. None of the hyper-parameters or network architectures are altered from the original SAC \cite{haarnoja2018soft}, PPO \cite{schulman2017ppo}, PEBBLE \cite{lee2021pebble} nor PrefPPO \cite{lee2021pebble} algorithms. The batch size for training on the preference dataset is $M$, matching the number of queries presented to the teacher, and varies based on the amount of feedback. For details about model architectures and hyper-parameters, refer to Appendices \ref{app_sec:architecture_details} and \ref{app_sec:hyper_parameter_details}. All experiments were run on a single GPU (Tesla V100) with 10 CPU cores (Intel Xeon Gold). The policy and preference learning implementations provided in the B-Pref repository \footnote{\url{https://github.com/rll-research/BPref}} are used for all experiments.

\subsection{Joint Policy and Reward Learning} \label{sec:joint_results}

\begin{figure*}
\begin{center}
\includegraphics[width = 0.6\linewidth]{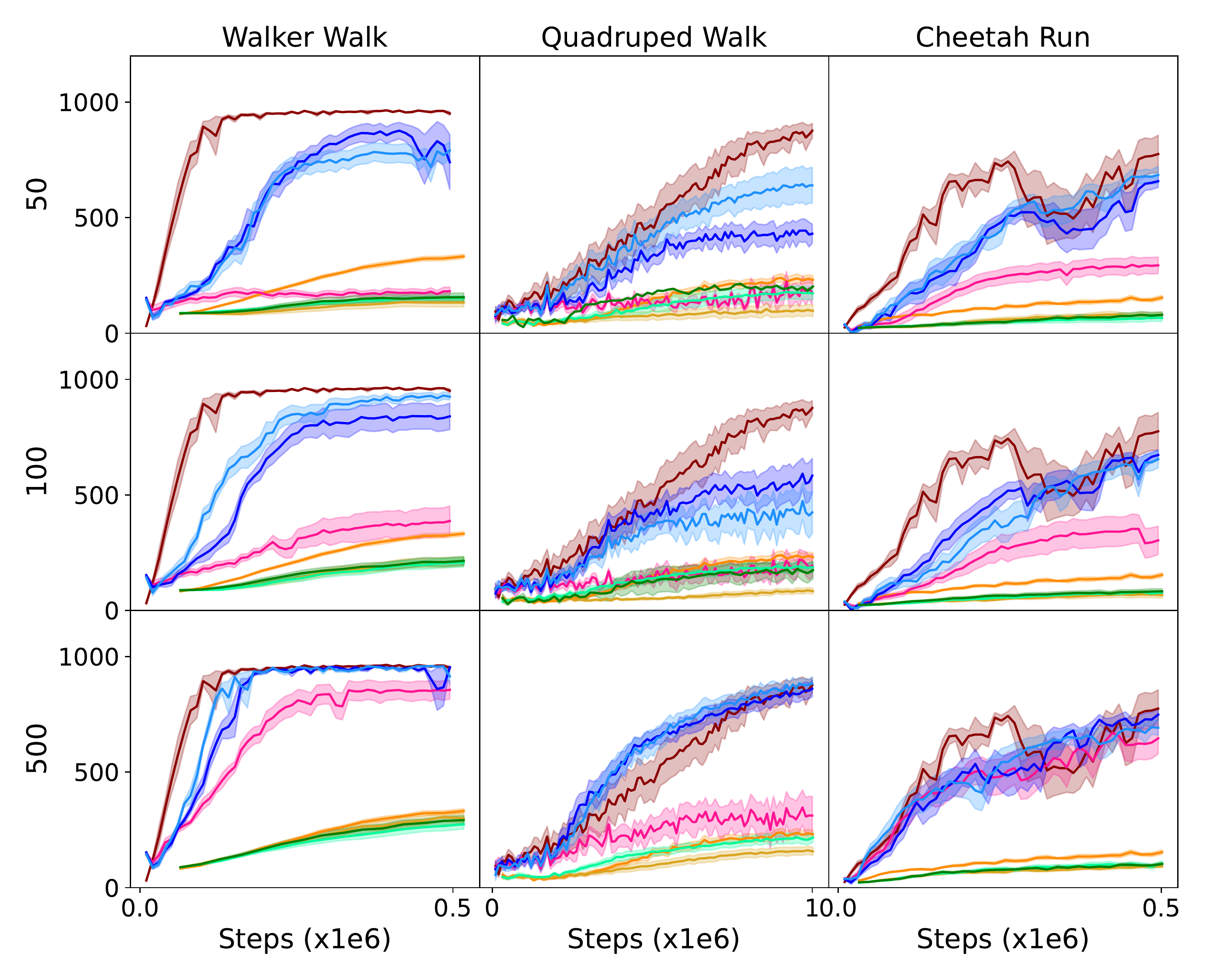}\hfill
\includegraphics[width = 0.4\linewidth]{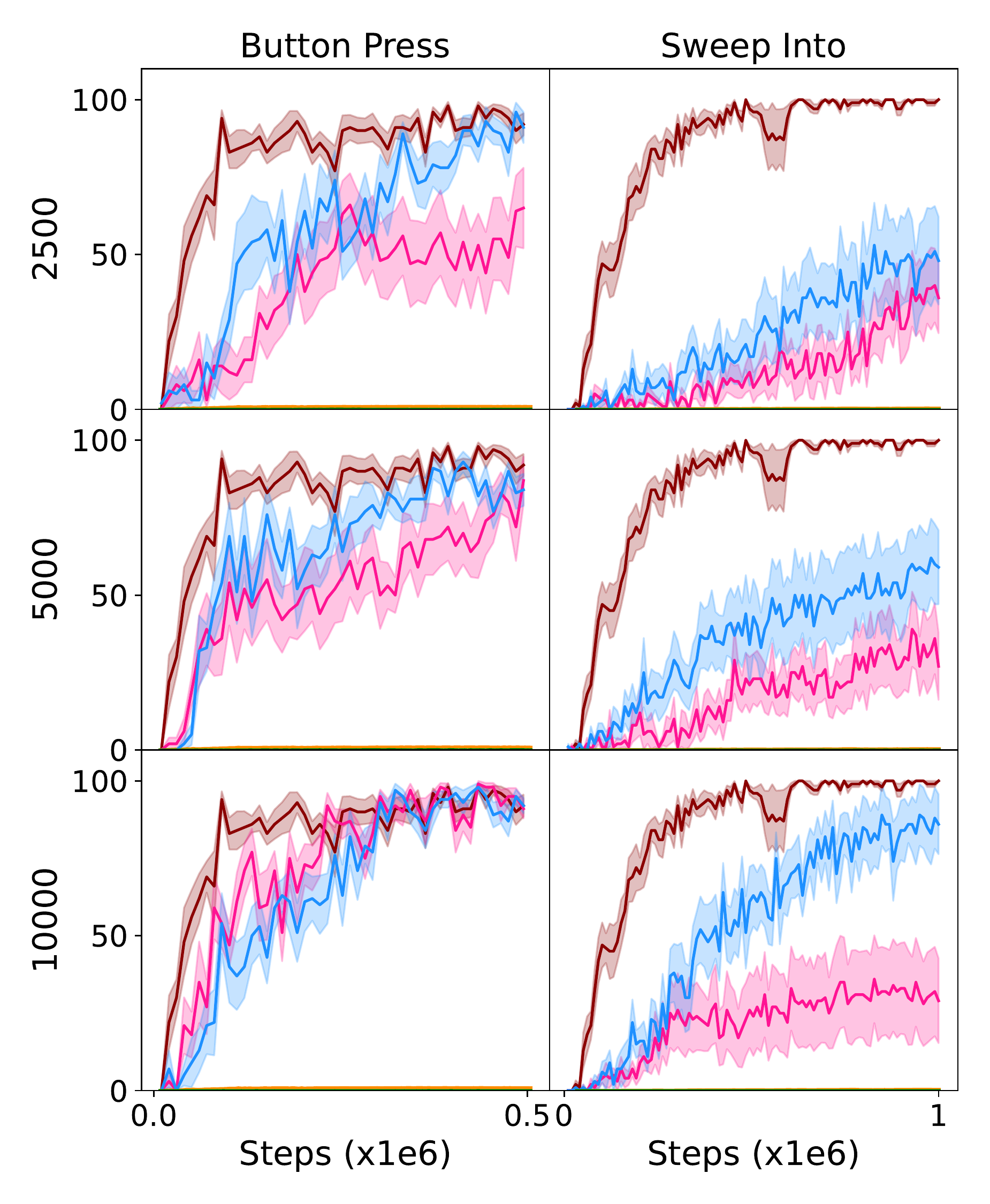}\hfill
\\[\smallskipamount]
\includegraphics[width = \linewidth]{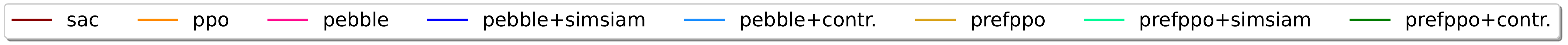}
\caption{Learning curves for walker-walk, quadruped-walk, cheetah-run, sweep into, and button press with 50, 100, \& 500 (walker, quadruped, \& cheetah) and 2.5k, 5k, \& 10k (sweep into \& button press) teacher-labelled queries with disagreement-based sampling and the oracle labelling strategy.}
\label{fig:learning_curves_subset_summary}
\end{center}
\end{figure*}

\begin{figure*}[ht] 
\centering
\subfloat{\includegraphics[width = 0.9\linewidth]{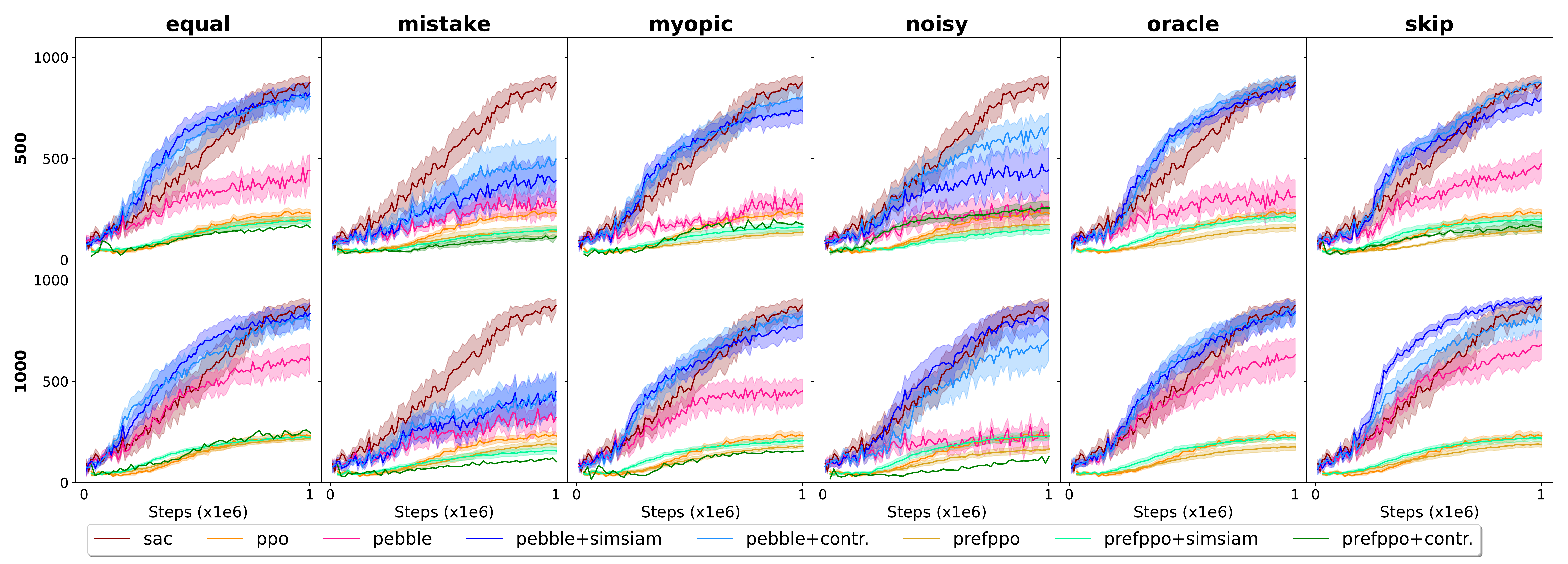}} \\
\caption{Impact of teacher labelling strategy on quadruped-walk with 500 \& 1k pieces of feedback.}
\label{fig:teacher_strategy_subset_summary}
\end{figure*}

\begin{figure*}[ht] 
\centering
\includegraphics[width = 0.49\linewidth]{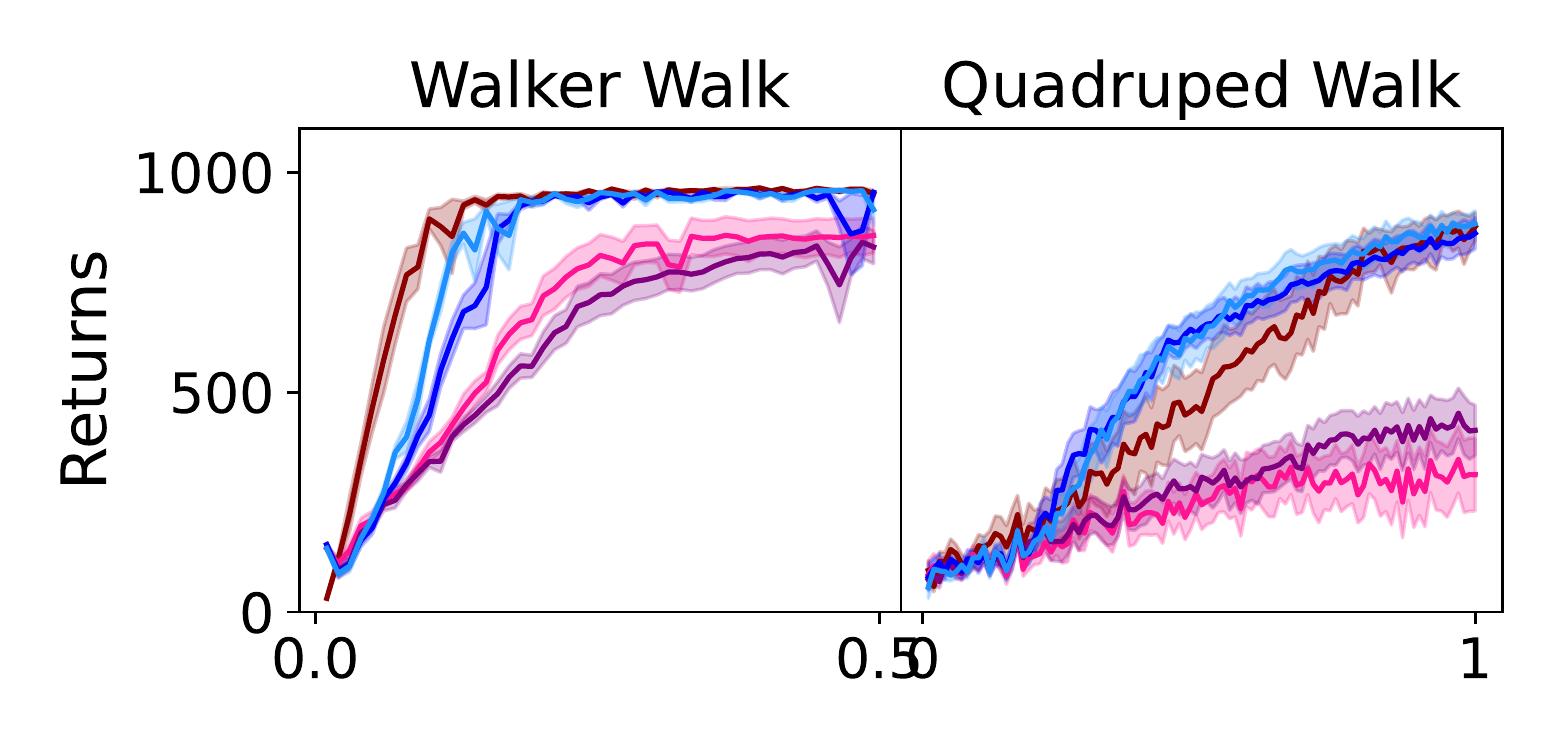} \hfill
\includegraphics[width = 0.49\linewidth]{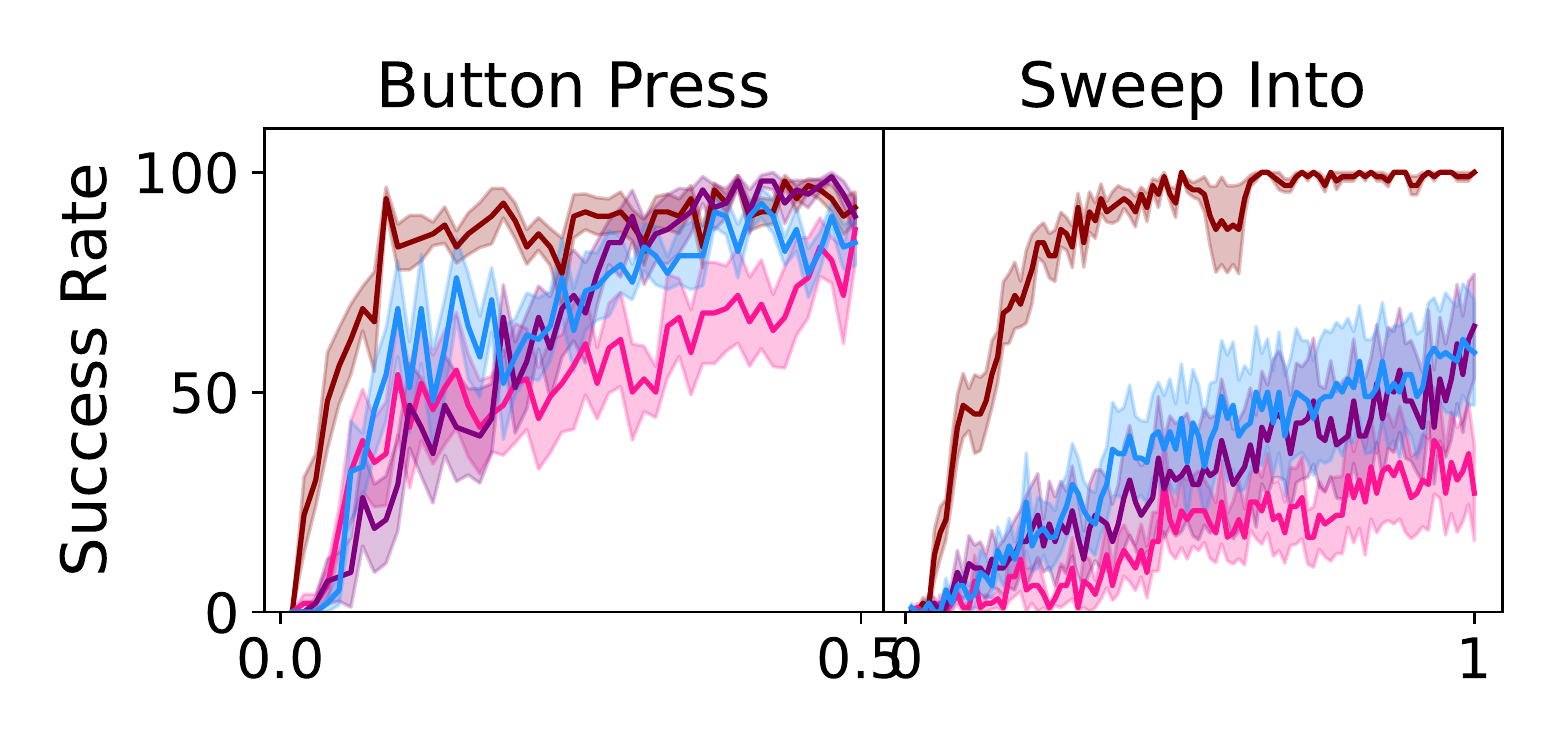} \hfill
\\[\smallskipamount]
\includegraphics[width = \linewidth]{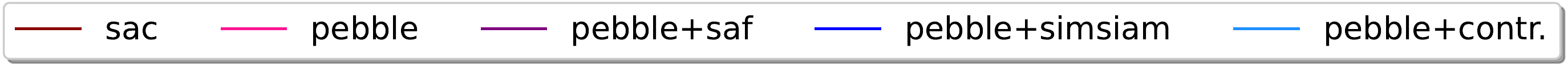} 
\caption{Ablation of the SAF reward net for walker-walk, quadruped-walk, sweep into, and button press with 500 (walker \& quadruped) and 5k (sweep into \& button press) teacher-labelled queries with disagreement-based sampling and the oracle labelling strategy.}
\label{fig:saf_ablation_subset_summary}
\end{figure*}

We follow the experiments outlined by the B-Pref benchmark \cite{lee2021bpref} and present results for SAC and PPO trained on the ground truth reward, PEBBLE, PrefPPO, PEBBLE + REED (SimSiam and Contrastive), and PrefPPO + REED (SimSiam and Contrastive); the REED conditions use the SAF reward network. Policy performance is evaluated with the ground truth reward function and is reported as mean and standard deviation over 10 runs. Learning curves are shown in Figure \ref{fig:learning_curves_subset_summary} and mean normalized returns \cite{lee2021bpref} in Table \ref{tab:perf_ratio_summary}, where mean normalized returns are given by:
\begin{equation}\label{eq:normalized_returns}
    \text{normalized returns} = \frac{1}{T}\sum_{t}{\frac{r_{\psi}(s_{t}, \pi^{\hat{r}_{\psi}}_{\phi}(a_{t}))}{r_{\psi}(s_{t}, \pi^{r_{\psi}}_{\phi}(a_{t}))}},
\end{equation}
where $T$ is the number of policy training training steps or episodes, $r_{\psi}$ is the ground truth reward function, $\pi^{\hat{r}_{\psi}}_{\phi}$ is the policy trained on the learned reward function, and $\pi^{r_{\psi}}_{\phi}$ is the policy trained on the ground truth reward function. 

The learning curves in Figures \ref{fig:learning_curves_subset_summary} and \ref{fig:teacher_strategy_subset_summary} show that REED significantly improves the speed of policy learning and the final performance of the learned policy relative to non-REED methods. This increase in policy performance is observed across environments, labelling strategies, and amounts of feedback. There are no clear benefits to SimSiam over contrastive REED objectives, which suggests the improvements in policy performance stem from encoding environment dynamics rather than any particular self-supervised objective. The benefits of encoding environment dynamics are especially pronounced for labelling types that introduce incorrect labels (i.e.\ mistake and noisy) and for smaller amounts of preference feedback. For example, on DMC tasks with 50 pieces of teacher feedback, REED methods more closely recover the performance of the policy trained on the ground truth reward recovering 62 -- 66\% vs.\ 21\% on walker-walk, and 65 -- 85\% vs.\ 38\% on quadruped-walk for PEBBLE-based methods (Table \ref{tab:perf_ratio_summary}).  We ablated the impact of the modified reward architecture and found that the performance improvements are not due to the modified reward network architecture (Figure \ref{fig:saf_ablation_subset_summary}). On Meta-World tasks, policy improvements are smaller for REED reward functions than they are for DMC. In particular, SimSiam-based REED methods frequently suffer representation collapse and are not reported here. We hypothesize this is due to how the states are specified. All MetaWorld states have a cosine similarity close to $1.0$, resulting in a trivial solution that just applies the identity to the states for the SimSiam objective. Using the contrastive objective improves MetaWorld performance, the performance gap is small because attending to goal state features (which differ across episodes) trivializes the problem. We would expect to see larger benefits when evaluating on the image-based MetaWorld tasks.

The results reported here demonstrate that encoding environment dynamics in the reward function results in better policy learning in preference-based RL.

\section{Discussion}\label{sec:discussion}

REED improves policy performance for preference-based RL by encoding environment dynamics for joint policy and reward learning (Section \ref{sec:joint_results}), and reward reuse (Section \ref{sec:joint_results}). 

The benefits are most pronounced for small amounts of preference feedback (\(\leq 500\)) and labelling styles that introduce errors into the preference labels.  Our stability results suggest that REED decreases the variance in reward predictions across reward updates and the benefits of lowering reward model variance are especially clear in the on-policy PrefPPO performance (Section \ref{sec:joint_results}).

Reducing the amount of feedback required for a person to change the behavior of an agent is an important step towards empowering people to align the agent behaviors with their needs. However, as with all technologies that aim to adapt and learn online from human feedback, there is the potential downside that people may encourage or teach an agent to behave in a way that is ultimately harmful to themselves and/or others. Therefore, it is important for guard rails to be deployed with \emph{all systems} that adapt based on human feedback, such as social rules and laws that cannot be violated or ensuring agent behaviors are always compliant with population level norms and rules.

\section{Conclusion \& Next Steps}

We have extended preference-based RL with a self-supervised temporal consistency task resulting in improved policy training and performance, \textit{specially for smaller amounts of feedback}. This work is an important step towards improving sample efficiency, which is necessary to make learning reward functions aligned with user preferences practical. 

Next steps include exploring the stability and generalizability of REED compared to the PEBBLE, improving the temporal consistency task on the MetaWorld environment, evaluating the benefits of the temporal consistency task on image-based domains, and conducting a user study to understand how the inclusion of REED improves the experience of human teachers.

\section*{References}
\medskip
\bibliographystyle{unsrtnat}
\bibliography{main.bib}

\appendix

\section{Appendix}

\subsection{Disagreement Sampling}\label{app_sec:disagreement_sampling}
 For all experiments in this paper, disagreement sampling is used to select which trajectory pairs will be presented to the teacher for preference labels. Disagreement-based sampling selects trajectory pairs as follows: (1) $N$ segments are sampled uniformly from the replay buffer; (2) the $M$ pairs of segments with the largest variance in preference prediction across the reward network ensemble are sub-sampled. Disagreement-based sampling is used as it reliably resulted in highest performing policies compared to the other sampling methods discussed in \citet{lee2021bpref}.
 
\subsection{Labelling Strategies}\label{app_sec:labelling_strategies}

An overview of the six labelling strategies is provided below, ordered from least to most noisy (see \cite{lee2021bpref} for details and configuration specifics):

\begin{enumerate}
    \item \textbf{oracle} - prefers the trajectory segment with the larger return and equally prefers both segments when their returns are identical
    \item \textbf{skip} - follows oracle, except randomly selects $10\%$ of the $M$ query pairs to discard from the preference dataset \(\mathcal{D}_{pref}\)
    \item \textbf{myopic} - follows oracle, except compares discounted returns ($\gamma=0.9$) placing more weight on transitions at the end of the trajectory
    \item \textbf{equal} - follows oracle, except marks trajectory segments as equally preferable when the difference in returns is less than $0.5\%$ of the average ground truth returns observed during the last $K$ policy training steps 
    \item \textbf{mistake} - follows oracle, except randomly selects $10\%$ of the $M$ query pairs and assigns incorrect labels in a structured way (e.g., a preference for segment two becomes a preference for segment one)
    \item \textbf{noisy} - randomly assigns labels with probability proportional to the relative returns associated with the pair, but labels the segments as equally preferred when they have identical returns 
\end{enumerate}
\vfill
\pagebreak
\subsection{REED Algorithm}\label{app_sec:pebble_pretrain_sfc_algorithm}

The REED task is specified in Algorithm \ref{alg:training_loop} in the context of the PEBBLE preference-learning algorithm. The main components of the PEBBLE algorithm are included, with our modifications identified in the comments. For the original and complete specification of PEBBLE, please see \cite{lee2021pebble} - Algorithm 2.

\algrenewcommand\algorithmicloop{}

\begin{algorithm}[h!]
\caption{PEBBLE + REED Training Procedure}
\label{alg:training_loop}
\begin{algorithmic}[1]
\Loop {\bfseries Given:}
  \State $K$
  \Comment{teacher feedback frequency}
  \State $M$
  \Comment{queries per feedback}
\EndLoop
\Loop {\bfseries Initializes:}
  \State $Q_{\theta}$
  \Comment{parameters for Q-function}
  \State $\hat{r}_{\psi}$
  \Comment{learned reward function}
  \State $\text{SPR}_{(\psi,\theta)}$
  \Comment{self-future consistency ($\psi$ parameters shared with $\hat{r}_{\psi}$)}
  \State $\mathcal{D}_{\text{pref}} \gets \emptyset$
  \Comment{preference dataset}
  \State $\mathcal{D}_{\text{SPR}} \gets \emptyset$
  \Comment{SPR dataset}
\EndLoop

\Statex
\LComment{unsupervised policy training and exploration}
\State $\mathcal{B}$, $\pi_{\phi} \gets \text{EXPLORE}()$
\Comment{\cite{lee2021pebble} - Algorithm 1}

\Statex
\LComment{joint policy and reward training}
\For{\text{policy train step}}
\If{step \% $K = 0$}
  \State $D_{\text{sfc}} \gets D_{\text{sfc}} \bigcup \mathcal{B}$
  \Comment{update SPR dataset}
  \For{each SPR gradient step}
      \State $\{(s_{t}, a_{t}, s_{t+1})\} \sim \mathcal{D}_{\text{sfc}}$
      \Comment{sample minibatch}
\State $\{(\hat{z}^{s}_{t+1}, z^{s}_{t+1})\} \gets \text{SFC\_FORWARD}(\{(s_{t}, a_{t}, s_{t+1})\})$
\Comment{Section \ref{func:sfc_forward}}
\State \textbf{optimize} $\mathcal{L}^{\text{reed}}$ with respect to $\text{SPR}_{(\psi,\theta)}$
\Comment{Equations (\ref{eq:simsiam_objective}) and (\ref{eq:contrastive_objective})}
\EndFor
\State $\hat{r}_{\psi} \gets \text{SPR}_{\psi}$
\Comment{copy shared SPR parameters to reward model}  
\State \textbf{update} $\mathcal{D}_{\text{pref}}$, $\hat{r}_{\psi}$, and $\mathcal{B}$
\Comment{following \cite{lee2021pebble} - Algorithm 2 [lines 9 - 18]}
\EndIf
\State \textbf{update} $\mathcal{B}$, $\pi_{\phi}$, and $Q_{\theta}$
\Comment {following \cite{lee2021pebble} - Algorithm 2 [lines 20 - 27]}
\EndFor
\end{algorithmic}
\end{algorithm}

\clearpage
\subsection{Architectures}\label{app_sec:architecture_details}

The network architectures are specified in PyTorch. For architecture hyper-parameters, e.g.\ hidden size and number of hidden layers, see Section \ref{app_sec:architecture_hyperparameter}

\subsubsection{Self-Predictive Representations Network} \label{app_sec:sfc_architecture}

The SPR network is implemented in PyTorch and is initialized as follows:

\definecolor{NiceGray}{cmyk}{0.3, 0.2, 0.2, 0.6}
\definecolor{NiceBlue}{cmyk}{1.0, 0.61, 0.0, 0.35}
\definecolor{NiceRed}{cmyk}{0.0, 0.85, 0.85, 0.22}
\lstset{
    language=Python,
    keywordstyle=\color{NiceBlue}\bfseries,
    basicstyle=\ttfamily\footnotesize,
    commentstyle=\color{NiceGray}\ttfamily,
    stringstyle=\color{NiceRed},
    numbers=none,
    stepnumber=5,
    numbersep=8pt,
    showstringspaces=false,
    breaklines=false,
    frame=none,
    tabsize=2,
    captionpos=b,
    morekeywords={self, with, True, False},
    mathescape=true,
    escapebegin=\color{NiceGray},
}

\begin{lstlisting}[]
def build_spr_network(
    self,
    state_size: int,
    state_embed_size: int,
    action_size: int,
    action_embed_size: int,
    hidden_size: int,
    consistency_comparison_dim: int,
    num_layers: int,
):
    """
    The network architecture and build logic to the complete the REED 
    self-supervised temporal consistency task based on SPR.
    
    Args:
      state_size: number of features defining the agent's state space
      state_embed_size: number of dimensions in the state embedding
      action_size: number of features defining the agent's actions
      action_embed_size: number of dimensions in the action embedding
      hidden_size: number of dimensions in the hidden layers of
                   state-action embedding network
      consistency_comparison_dim: number of units used to compare the
                                  predicted and target latent next 
                                  state
      num_layers: number of hidden layers used to embed the 
                  state-action representation
    """
    # build the network that will encode the state features
    self.state_encoder = torch.nn.Sequential(
        OrderedDict(
            [
                (
                    "state_dense1",
                    torch.nn.Linear(state_size, state_embed_size),
                ),
                (
                    "state_leakyrelu1",
                    torch.nn.LeakyReLU(negative_slope=1e-2),
                ),
            ]
        )
    )

    # build the network that will encode the action features
    self.action_encoder = torch.nn.Sequential(
        OrderedDict(
            [
                (
                    "action_dense1",
                    torch.nn.Linear(action_size, action_embed_size),
                ),
                (
                    "action_leakyrelu1",
                    torch.nn.LeakyReLU(negative_slope=1e-2),
                ),
            ]
        )
    )

    # build the network that models the relationship between the
    # state and action embeddings
    state_action_encoder = []
    hidden_in_size = action_embed_size + state_embed_size
    for i in range(num_layers):
        state_action_encoder.append(
            (
                f"trunk_dense{i+1}",
                torch.nn.Linear(hidden_in_size, hidden_size),
            )
        )
        state_action_encoder.append(
            (
                f"trunk_leakyrelu{i+1}",
                torch.nn.LeakyReLU(negative_slope=1e-2),
            )
        )
        hidden_in_size = hidden_size
    self.state_action_encoder = torch.nn.Sequential(
        OrderedDict(state_action_encoder)
    )

    # this is a single dense layer because we want to focus as much of
    # the useful semantic information as possible in the state-action
    # representation
    self.next_state_predictor = torch.nn.Linear(
        hidden_size, state_embed_size
    )

    self.next_state_projector = torch.nn.Linear(
        state_embed_size, state_embed_size
    )

    self.consistency_predictor = nn.Linear(
        consistency_comparison_dim, consistency_comparison_dim
    )
\end{lstlisting}

A forward pass through the SFC network is as follows:

\begin{lstlisting}[label={func:sfc_forward}]
def spr_forward(self, transitions: EnvironmentTransitionBatch):
    """
    The logic for a forward pass through the SPR network.
    Args:
      transitions: a batch of environment transitions composed of
                   states, actions, and next states
    Returns:
      predicted embedding of the next state - p in SimSiam paper
      next state embedding (detached from graph) - z in SimSiam paper
      dimensionality: (batch, time step)
    """
    # encode the state, the action, and the state-action pair
    # $s_{t} \rightarrow z^{s}_{t}$
    states_embed = self.state_encoder(transitions.states)
    # $a_{t} \rightarrow z^{a}_{t}$
    actions_embed = self.action_encoder(transitions.actions)
    # $(s_{t}, a_{t}) \rightarrow z^{sa}_{t}$
    state_action_embeds = torch.concat(
        [states_embed, actions_embed], dim=-1
    )
    state_action_embed = self.state_action_encoder(
        state_action_embeds
    )

    # predict and project the representation of the next state
    # $z^{sa}_{t} \rightarrow \hat{z}^{s}_{t+1}$
    next_state_pred = self.next_state_predictor(state_action_embed)
    next_state_pred = self.next_state_projector(next_state_pred)
    next_state_pred = self.consistency_predictor(next_state_pred)

    # we don't want gradients to back-propagate into the learned
    # parameters from anything we do with the next state
    with torch.no_grad():
        # $s_{t+1} \rightarrow z^{s}_{t+1}$
        # embed the next state
        next_state_embed = self.state_encoder(transitions.next_states)
        # project the next state embedding into a space where it can be
        # compared with the predicted next state
        projected_next_state_embed = self.next_state_projector(
            next_state_embed
        )

    # from the SimSiam paper, this is p and z
    return next_state_pred, projected_next_state_embed
\end{lstlisting}

\subsubsection{SAF Reward Network}

The architecture of the SAF Reward Network is a subset of the SFC network with the addition of a linear to map the state-action representation to predicted rewards. The SFC network is implemented in PyTorch and is initialized following the below build method:

\begin{lstlisting}[]
def build_saf_network(
    self,
    state_size: int,
    state_embed_size: int,
    action_size: int,
    action_embed_size: int,
    hidden_size: int,
    num_layers: int,
    final_activation_type: str,
):
    """
    Args:
      state_size: number of features defining the agent's state space
      state_embed_size: number of dimensions in the state embedding
      action_size: number of features defining the agent's actions
      action_embed_size: number of dimensions in the action embedding
      hidden_size: number of dimensions in the hidden layers of
                   state-action embedding network
      num_layers: number of hidden layers used to embed the 
                  state-action representation
      final_activation_type: the activation used on the final layer
    """
    # build the network that will encode the state features
    self.state_encoder = torch.nn.Sequential(
        OrderedDict(
            [
                (
                    "state_dense1",
                    torch.nn.Linear(state_size, state_embed_size),
                ),
                (
                    "state_leakyrelu1",
                    torch.nn.LeakyReLU(negative_slope=1e-2),
                ),
            ]
        )
    )

    # build the network that will encode the action features
    self.action_encoder = torch.nn.Sequential(
        OrderedDict(
            [
                (
                    "action_dense1",
                    torch.nn.Linear(action_size, action_embed_size),
                ),
                (
                    "action_leakyrelu1",
                    torch.nn.LeakyReLU(negative_slope=1e-2),
                ),
            ]
        )
    )

    # build the network that models the relationship between the state
    # and action embeddings
    state_action_encoder = []
    hidden_in_size = action_embed_size + state_embed_size
    for i in range(num_layers):
        state_action_encoder.append(
            (
                f"trunk_dense{i+1}",
                torch.nn.Linear(hidden_in_size, hidden_size),
            )
        )
        state_action_encoder.append(
            (
                f"trunk_leakyrelu{i+1}",
                torch.nn.LeakyReLU(negative_slope=1e-2),
            )
        )
        hidden_in_size = hidden_size
    self.state_action_encoder = torch.nn.Sequential(
        OrderedDict(state_action_encoder)
    )

    # build the prediction head and select a final activation
    self.prediction_head = torch.nn.Linear(hidden_size, 1)
    if final_activation_type == "tanh":
        self.final_activation = torch.nn.Tanh()
    elif final_activation_type == "sig":
        self.final_activation = torch.nn.Sigmoid()
    else:
        self.final_activation_type = torch.nn.ReLU()
\end{lstlisting}

A forward pass through the SAF network is as follows:

\begin{lstlisting}[label={func:sfc_forward}]
def saf_forward(self, transitions: EnvironmentTransitionBatch):
    """
    Args:
      transitions: a batch of environment transitions composed of
                   states, actions, and next states
    Returns:
      predicted embedding of the next state - p in SimSiam paper
      next state embedding (detached from graph) - z in SimSiam paper
      dimensionality: (batch, time step)
    """
    # encode the state, the action, and the state-action pair
    # $s_{t} \rightarrow z^{s}_{t}$
    states_embed = self.state_encoder(transitions.states)
    # $a_{t} \rightarrow z^{a}_{t}$
    actions_embed = self.action_encoder(transitions.actions)
    # $(s_{t}, a_{t}) \rightarrow z^{sa}_{t}$
    state_action_embeds = torch.concat(
        [states_embed, actions_embed], dim=-1
    )
    state_action_embed = self.state_action_encoder(
        state_action_embeds
    )

    return self.final_activation(
        self.prediction_head(state_action_embed)
    )
\end{lstlisting}

\subsection{Hyper-parameters}\label{app_sec:hyper_parameter_details}

\subsubsection{Train Hyper-parameters}\label{app_sec:train_hyperparameter}

This section specifies the hyper-parameters (e.g.\ learning rate, batch size, etc) used for the experiments and results (Section \ref{sec:experiments}). The SAC, PPO, PEBBLE, and PrefPPO experiments all match those used in \cite{haarnoja2018soft}, \cite{schulman2017ppo}, and \cite{lee2021pebble} respectively. The SAC and PPO hyper-parameters are specified in Table \ref{app_tab:sac_ppo_train_hyperparameters}, the PEBBLE and PrefPPO hyper-parameters are give in Table \ref{app_tab:pebble_prefppo_train_hyperparameters}, and the hyper-parameters used to train on the REED task are in Table \ref{app_tab:reed_train_hyperparameters}.

\begin{table}[h]
\centering
\begin{footnotesize}
\caption{Training hyper-parameters for SAC \cite{haarnoja2018soft} and PPO \cite{schulman2017ppo}.}
\renewcommand{\arraystretch}{1.6}
\label{app_tab:sac_ppo_train_hyperparameters}
\begin{tabular}{c m{0.41\textwidth} m{0.52\textwidth}}
    \toprule
    & \multicolumn{1}{c}{\textsc{Hyper-parameter}} & \multicolumn{1}{c}{\textsc{Value}} \\
    \midrule
    \multicolumn{3}{c}{\textsc{SAC}} \\
    \midrule
    & Learning rate & 1e-3 (cheetah), 5e-4 (walker), \newline 1e-4 (quadruped), 3e-4 (MetaWorld) \\
    & Batch size & \(512\) (DMC), \(1024\) (MetaWorld) \\
    & Total timesteps & \(500\)k (cheetah, walker, button press), \newline \(1\)M (quadruped, sweep into) \\
    & Optimizer &Adam \cite{kingma2015adam} \\
    & Critic EMA \(\tau\) & \(5e-3\) \\
    & Critic target update freq. & \(2\) \\
    & (\(\mathcal{B}_{1}, \mathcal{B}_{2}\)) & \((0.9, 0.999)\) \\
    & Initial Temperature & \(0.1\) \\
    & Discount \(\gamma\) & \(0.99\) \\
    \midrule
    \multicolumn{3}{c}{\textsc{PPO}}\\
    \midrule
    & Learning rate & 5e-5 (DMC), 3e-4 (MetaWorld)\\
    & Batch size & \(128\) (all but cheetah), \(512\) (cheetah) \\
    & Total timesteps & \(500\)k (cheetah, walker, button press), \newline \(1\)M (quadruped, sweep into) \\
    & Envs per worker & \(8\) (sweep into), \(16\) (cheetah, quadruped), \newline \(32\) (walker, sweep into) \\
    & Optimizer &Adam \cite{kingma2015adam} \\
    & Discount \(\gamma\) & \(0.99\) \\
    & Clip range & \(0.2\) \\
    & Entropy bonus & \(0.0\) \\
    & GAE parameter \(\lambda\) & \(0.92\) \\
    & Timesteps per rollout & \(250\) (MetaWorld), \(500\) (DMC) \\
    \bottomrule
\end{tabular}
\end{footnotesize}
\end{table}

\begin{table}[h]
    \centering
    \begin{footnotesize}
    \caption{Training hyper-parameters for PEBBLE \cite{lee2021pebble} and PrefPPO \cite{christiano2017deep,lee2021pebble}. The only hyper-parameter that differs between PEBBLE and PrefPPO is the DMC learning rate. The batch size for the reward network changes based per total feedback amount to match the number of queries \(M\) sent to the teacher for labelling each feedback session.}
    \label{app_tab:pebble_prefppo_train_hyperparameters}
    \renewcommand{\arraystretch}{1.6}
    \begin{tabular}{m{0.41\textwidth} m{0.52\textwidth}}
    \toprule
        \textsc{Hyper-parameter} & \textsc{Value} \\
        \midrule
        Learning rate PEBBLE & 3e-4 \\
        Learning rate PrefPPO & 5e-4 (DMC), 3e-4 (MetaWorld) \\
        Optimizer & Adam \cite{kingma2015adam} \\
        Segment length \(l\) & \(50\) (DMC), \(25\) (MetaWorld) \\
        Feedback amount / number queries (\(M\)) & \(1\)k/\(100\), \(500\)/\(50\), \(200\)/\(20\), \(100\)/\(10\), \(50\)/\(5\) (DMC) \newline \(20\)k/\(100\), \(10\)k/\(50\), \(5\)k/\(25\), \(2.5\)k/\(12\) (MetaWorld) \\
        Steps between queries (\(K\)) & \(20\)k (walker, cheetah), \(30\)k (quadruped), \newline \(5\)k (MetaWorld)\\
    \bottomrule
    \end{tabular}
    \end{footnotesize}
\end{table}

\begin{table}[h]
    \centering
    \caption{Training hyper-parameters for REED with the SPR objective \cite{schwarzer2020data} (Section \ref{sec:encoding_env_dynamics}). The REED hyper-parameters were used with both the PEBBLE \cite{lee2021pebble} and PrefPPO \cite{christiano2017deep,lee2021pebble} preference-learning algorithms. Hyper-parameters are by environment/task and shared by the two SSL objectives: SimSiam vs. Contrastive (Section \ref{sec:encoding_env_dynamics}). Training on the REED task occurred every \(K\) steps (specified in Table \ref{app_tab:pebble_prefppo_train_hyperparameters}) prior to updating on the preference task. The SPR objective predicts future latent states \(k\) steps in the future. While our hyper-parameter sweep evaluated multiple values for \(k\), we found that \(k=1\) vs. \(k>1\) had no real impact on learning quality for these state-action feature spaces.}
    \label{app_tab:reed_train_hyperparameters}
    \begin{tabular}{lllll}
    \toprule
    \textsc{Environment} & \textsc{Learning Rate} & \textsc{Epochs per Update} & \textsc{Batch Size} & \textsc{Optimizer} \\
    \midrule
        Walker &  1e-3 & 20 & 12 & SGD \\
        Cheetah & 1e-3 & 20 & 12 & SGD \\
        Quadruped & 1e-4 & 20 & 128 & Adam \cite{kingma2015adam} \\
        Button Press & 1e-4 & 10 & 128 & Adam \cite{kingma2015adam} \\
        Sweep Into & 5e-5 & 5 & 256 & Adam \cite{kingma2015adam}\\
    \bottomrule
    \end{tabular}
\end{table}

\clearpage

\subsubsection{Architecture Hyper-parameters}\label{app_sec:architecture_hyperparameter}

The network hyper-parameters (e.g.\ hidden dimension, number of hidden layers, etc) used for the experiments and results (Section \ref{sec:experiments}) are specified in Table \ref{app_tab:architecture_hyperparameters}. 

\begin{table}[h]
    \centering
    \begin{footnotesize}
    \caption{Architecture hyper-parameters for SAC \cite{haarnoja2018soft}, PPO \cite{schulman2017ppo}, the base reward model (used for PEBBLE \cite{lee2021pebble} and PrePPO \cite{christiano2017deep,lee2021pebble}), the SAF reward model (Section \ref{sec:encoding_env_dynamics}), and the SPR model (Section \ref{sec:encoding_env_dynamics}). The hyper-parameters reported here are intended to inform the values to used to initialize the architectures in Section \ref{app_sec:architecture_details}. Hyper-parameters not relevant to a model are indicated with ``N/A''. The SPR model is what REED uses to construct the self-supervised temporal consistency task. The base reward model is  used with PEBBLE and PrefPPO in \citet{lee2021pebble} and \citep{lee2021bpref}. The SAF reward network is used for all REED conditions in Section \ref{sec:experiments}. The ``Final Activation'' refers to the activation function used just prior to predicting the reward for a given state action pair}
    \label{app_tab:architecture_hyperparameters}
    \hspace*{-0.75cm}
    \begin{tabular}{llllll}
    \toprule
        \textsc{Hyper-Parameter} & \textsc{SAC} & \textsc{PPO} & \textsc{Base Reward} & \textsc{SAF Reward} & \textsc{SPR Net} \\
        \midrule
        \multirow{4}{*}{State embed size} & \multirow{4}{*}{N/A} & \multirow{4}{*}{N/A} & \multirow{4}{*}{N/A} & \(20\) (walker), & \(20\) (walker), \\
         &  & & & \(17\) (cheetah), & \(17\) (cheetah), \\
         &  & & & \(78\) (quadruped), & \(78\) (quadruped), \\
         &  & & & \(30\) (Metaworld) & \(30\) (Metaworld) \\
        \midrule
        \multirow{4}{*}{Action embed size} & \multirow{4}{*}{N/A} & \multirow{4}{*}{N/A} &\multirow{4}{*}{N/A}  & \(10\) (walker), & \(10\) (walker), \\
         &  & & & \(6\) (cheetah), & \(6\) (cheetah), \\
         &  & & & \(12\) (quadruped), & \(12\) (quadruped), \\
         &  & & & \(4\) (Metaworld) & \(4\) (Metaworld) \\
        \midrule
        \multirow{4}{*}{Comparison units} & \multirow{4}{*}{N/A} & \multirow{4}{*}{N/A} & \multirow{4}{*}{N/A} & \(5\) (walker), & \(5\) (walker), \\
         &  & & & \(4\) (cheetah), & \(4\) (cheetah), \\
         &  & & & \(10\) (quadruped), & \(10\) (quadruped), \\
         &  & & & \(5\) (Metaworld) & \(5\) (Metaworld)\\
        \midrule
        \multirow{2}{*}{Num. hidden} & \(2\) (DMC),  & \multirow{2}{*}{3} & \multirow{2}{*}{3} & \multirow{2}{*}{3} & \multirow{2}{*}{3} \\
        & \(3\) (MetaWorld) & & & & \\
        \midrule
        \multirow{2}{*}{Units per layer} & \(1024\) (DMC),  & \multirow{2}{*}{256} & \multirow{2}{*}{256} & \multirow{2}{*}{256} & \multirow{2}{*}{256} \\
        & \(256\) (MetaWorld) & & & & \\
        \midrule
        Final activation & N/A & N/A & tanh & tanh & N/A \\
    \bottomrule
    \end{tabular}
    \end{footnotesize}
\end{table}

\subsection{SAF Reward Net Ablation}\label{app_sec:saf_ablation}

\begin{table*}[h]
\caption{The impact of the SAF reward network is ablated. Ratio of policy performance on learned versus ground truth rewards for \textbf{walker-walk}, \textbf{quadruped-walk}, \textbf{sweep into}, and \textbf{button press} across preference learning methods, labelling methods and feedback amounts (with disagreement sampling).}
\label{app_tab:completeperf_ratio_pebble_pebblesfc}
\begin{center}
\begin{scriptsize}
\begin{sc}
\hspace*{-1cm}
\setlength\tabcolsep{4.5pt} 
\begin{tabular}{lll @{\hspace{1cm}} cccccc @{\hspace{1cm}} c}
\toprule
& Feedback & Method & Oracle & Mistake & Equal & Skip & Myopic & Noisy & Mean  \\
\midrule
\multicolumn{10}{c}{Walker-walk} \\
\midrule
 & \multirow{4}{*}{1k} & PEBBLE & 0.85 (0.17) & 0.76 (0.21) & 0.88 (0.16) & 0.85 (0.17) & 0.79 (0.18) & 0.81 (0.18) & 0.83 \\ 
 & & \hspace{4 mm}+SAF & 0.81 (0.19) & 0.62 (0.18) & 0.88 (0.16) & 0.81 (0.19) & 0.74 (0.17) & 0.81 (0.19) & 0.78 \\
 & & \hspace{4 mm}+SimSiam  & 0.9 (0.16) & 0.77 (0.2) & 0.91 (0.12) & 0.89 (0.16) & 0.8 (0.17) & 0.88 (0.17) & 0.86 \\ 
 & & \hspace{4 mm}+Contr.  & 0.9 (0.16) & 0.77 (0.2) & 0.91 (0.12) & 0.89 (0.16) & 0.8 (0.17) & 0.88 (0.17) & 0.86 \\ 
 \cmidrule{2-10}
& \multirow{4}{*}{500} & PEBBLE & 0.74 (0.18) & 0.61 (0.17) & 0.84 (0.19) & 0.75 (0.19) & 0.67 (0.19) & 0.69 (0.19) & 0.72 \\
 & & \hspace{4 mm}+SAF & 0.68 (0.17) & 0.51 (0.13) & 0.76 (0.17) & 0.68 (0.17) & 0.56 (0.15) & 0.68 (0.17) & 0.65 \\
 & & \hspace{4 mm}+SimSiam & 0.86 (0.2) & 0.71 (0.2) & 0.87 (0.2) & 0.87 (0.2) & 0.82 (0.22) & 0.84 (0.2) & 0.83 \\ 
 & & \hspace{4 mm}+Contr. & 0.9 (0.17) & 0.81 (0.19) & 0.9 (0.14) & 0.9 (0.17) & 0.88 (0.16) & 0.88 (0.18) & 0.88 \\ 
 \cmidrule{2-10}
& \multirow{4}{*}{250} & PEBBLE & 0.59 (0.17) & 0.41 (0.12) & 0.67 (0.2) & 0.56 (0.17) & 0.43 (0.13) & 0.51 (0.13) & 0.53 \\ 
 & & \hspace{4 mm}+SAF & 0.53 (0.16) & 0.41 (0.15) & 0.59 (0.18) & 0.53 (0.16) & 0.36 (0.1) & 0.48 (0.14) & 0.48 \\
 & & \hspace{4 mm}+SimSiam & 0.8 (0.23) & 0.6 (0.16) & 0.85 (0.21) & 0.8 (0.24) & 0.75 (0.26) & 0.8 (0.24) & 0.77 \\
 & & \hspace{4 mm}+Contr. & 0.85 (0.19) & 0.73 (0.23) & 0.85 (0.19) & 0.85 (0.2) & 0.79 (0.2) & 0.85 (0.22) & 0.82 \\
\midrule
\multicolumn{10}{c}{Quadruped-Walk} \\
\midrule
& \multirow{4}{*}{2k} & PEBBLE & 0.94 (0.15) & 0.55 (0.19) & 1.1 (0.26) & 1.0 (0.16) & 0.93 (0.13) & 0.56 (0.19) & 0.86 \\ 
 & & \hspace{4 mm}+SAF & 0.97 (0.15) & 0.45 (0.17) & 1.2 (0.22) & 0.87 (0.19) & 0.76 (0.13) & 0.59 (0.14) & 0.81 \\ 
 & & \hspace{4 mm}+SimSiam & 1.3 (0.31) & 0.47 (0.19) & 1.4 (0.37) & 1.3 (0.26) & 1.2 (0.18) & 0.96 (0.15) & 1.09 \\ 
 & & \hspace{4 mm}+Contr. & 1.3 (0.25) & 0.7 (0.16) & 1.2 (0.24) & 1.3 (0.29) & 1.3 (0.28) & 1.0 (0.16) & 1.13 \\ 
 \cmidrule{2-10}
 & \multirow{4}{*}{1k} & PEBBLE & 0.86 (0.15) & 0.53 (0.19) & 0.88 (0.15) & 0.91 (0.14) & 0.73 (0.18) & 0.48 (0.25) & 0.73 \\ 
 & & \hspace{4 mm}+SAF & 0.79 (0.16) & 0.44 (0.19) & 0.99 (0.23) & 0.9 (0.19) & 0.63 (0.15) & 0.6 (0.2) & 0.72 \\ 
 & & \hspace{4 mm}+SimSiam & 1.1 (0.19) & 0.59 (0.14) & 1.2 (0.22) & 1.3 (0.3) & 1.1 (0.21) & 1.0 (0.15) & 1.04 \\ 
 & & \hspace{4 mm}+Contr. & 1.1 (0.19) & 0.63 (0.16) & 1.2 (0.29) & 1.1 (0.19) & 1.1 (0.19) & 0.83 (0.14) & 0.99 \\ 
 \cmidrule{2-10}
& \multirow{4}{*}{500} & PEBBLE & 0.56 (0.21) & 0.48 (0.21) & 0.66 (0.2) & 0.64 (0.15) & 0.47 (0.22) & 0.48 (0.23) & 0.55 \\ 
 & & \hspace{4 mm}+SAF & 0.63 (0.16) & 0.4 (0.22) & 0.85 (0.14) & 0.75 (0.19) & 0.56 (0.18) & 0.5 (0.19) & 0.61 \\
 & & \hspace{4 mm}+SimSiam & 1.1 (0.21) & 0.58 (0.16) & 1.2 (0.24) & 1.0 (0.22) & 1.0 (0.19) & 0.68 (0.16) & 0.93 \\
 & & \hspace{4 mm}+Contr.  & 1.1 (0.21) & 0.64 (0.11) & 1.1 (0.22) & 1.1 (0.17) & 1.0 (0.17) & 0.85 (0.14) & 0.97 \\ 
 \cmidrule{2-10}
& \multirow{4}{*}{250} & PEBBLE & 0.53 (0.18) & 0.36 (0.23) & 0.64 (0.15) & 0.62 (0.16) & 0.46 (0.22) & 0.47 (0.21) & 0.51 \\ 
 & & \hspace{4 mm}+SAF & 0.51 (0.2) & 0.36 (0.22) & 0.73 (0.18) & 0.53 (0.17) & 0.53 (0.19) & 0.45 (0.24) & 0.52 \\ 
 & & \hspace{4 mm}+SimSiam & 0.98 (0.15) & 0.58 (0.18) & 1.0 (0.19) & 0.79 (0.12) & 0.9 (0.18) & 0.77 (0.16) & 0.84 \\ 
 & & \hspace{4 mm}+Contr. & 0.98 (0.15) & 0.58 (0.18) & 1.0 (0.19) & 0.79 (0.12) & 0.9 (0.18) & 0.77 (0.16) & 0.84 \\
\midrule
\multicolumn{10}{c}{Button Press} \\
\midrule
& \multirow{3}{*}{20k} & PEBBLE & 0.72 (0.26) & 0.57 (0.26) & 0.77 (0.25) & 0.75 (0.26) & 0.68 (0.21) & 0.72 (0.24) & 0.70 \\ 
 &  & \hspace{4 mm}+SAF & 0.77 (0.23) & 0.72 (0.28) & 0.84 (0.23) & 0.75 (0.24) & 0.78 (0.21) & 0.77 (0.22) & 0.77 \\ 
 &  & \hspace{4 mm}+Contr. & 0.65 (0.25) & 0.61 (0.28) & 0.67 (0.27) & 0.67 (0.27) & 0.67 (0.24) & 0.69 (0.26) & 0.66 \\ 
 \cmidrule{2-10}
 & \multirow{3}{*}{10k} & PEBBLE & 0.66 (0.26) & 0.47 (0.21) & 0.67 (0.27) & 0.63 (0.26) & 0.67 (0.24) & 0.6 (0.26) & 0.62 \\ 
 &  & \hspace{4 mm}+SAF & 0.7 (0.25) & 0.66 (0.26) & 0.74 (0.23) & 0.71 (0.25) & 0.67 (0.19) & 0.71 (0.25) & 0.70 \\ 
 &  & \hspace{4 mm}+Contr. & 0.65 (0.27) & 0.61 (0.3) & 0.66 (0.27) & 0.62 (0.26) & 0.6 (0.25) & 0.68 (0.28) & 0.64 \\ 
 \cmidrule{2-10}
& \multirow{3}{*}{5k} & PEBBLE & 0.48 (0.21) & 0.31 (0.12) & 0.56 (0.25) & 0.54 (0.24) & 0.59 (0.23) & 0.52 (0.23) & 0.50 \\ 
 &  & \hspace{4 mm}+SAF & 0.63 (0.25) & 0.55 (0.24) & 0.65 (0.26) & 0.68 (0.24) & 0.62 (0.21) & 0.7 (0.24) & 0.64 \\ 
 &  & \hspace{4 mm}+Contr. & 0.55 (0.24) & 0.54 (0.26) & 0.65 (0.27) & 0.63 (0.26) & 0.57 (0.24) & 0.63 (0.28) & 0.60 \\ 
 \cmidrule{2-10}
& \multirow{3}{*}{2.5k} & PEBBLE & 0.37 (0.18) & 0.21 (0.088) & 0.44 (0.21) & 0.34 (0.15) & 0.4 (0.17) & 0.34 (0.18) & 0.35 \\ 
 &  & \hspace{4 mm}+SAF & 0.58 (0.26) & 0.38 (0.17) & 0.61 (0.26) & 0.54 (0.23) & 0.52 (0.21) & 0.54 (0.2) & 0.53 \\ 
 &  & \hspace{4 mm}+Contr. & 0.49 (0.25) & 0.42 (0.22) & 0.52 (0.24) & 0.5 (0.23) & 0.44 (0.17) & 0.45 (0.21) & 0.47 \\ 
\midrule
\multicolumn{10}{c}{Sweep Into} \\
\midrule
 & \multirow{3}{*}{20k} &  PEBBLE & 0.53 (0.25) & 0.26 (0.15) & 0.51 (0.23) & 0.52 (0.27) & 0.47 (0.28) & 0.47 (0.26) & 0.46 \\ 
 &  & \hspace{4 mm}+SAF & 0.5 (0.24) & 0.36 (0.15) & 0.47 (0.22) & 0.39 (0.19) & 0.49 (0.21) & 0.6 (0.21) & 0.47 \\ 
 &  & \hspace{4 mm}+Contr. & 0.5 (0.22) & 0.36 (0.13) & 0.41 (0.2) & 0.6 (0.22) & 0.54 (0.21) & 0.61 (0.25) & 0.50 \\ 
 \cmidrule{2-10}
 & \multirow{3}{*}{10k} & PEBBLE & 0.28 (0.12) & 0.22 (0.13) & 0.45 (0.21) & 0.33 (0.17) & 0.47 (0.25) & 0.51 (0.24) & 0.38 \\ 
 &  & \hspace{4 mm}+SAF & 0.41 (0.2) & 0.32 (0.19) & 0.48 (0.2) & 0.47 (0.17) & 0.46 (0.2) & 0.57 (0.24) & 0.45 \\ 
 &  & \hspace{4 mm}+Contr. & 0.47 (0.23) & 0.3 (0.14) & 0.45 (0.24) & 0.32 (0.21) & 0.42 (0.22) & 0.44 (0.21) & 0.40 \\ 
 \cmidrule{2-10}
& \multirow{3}{*}{5k} & PEBBLE & 0.17 (0.099) & 0.17 (0.089) & 0.28 (0.19) & 0.24 (0.15) & 0.23 (0.13) & 0.22 (0.12) & 0.22 \\ 
 &  & \hspace{4 mm}+SAF & 0.36 (0.15) & 0.2 (0.13) & 0.4 (0.23) & 0.38 (0.17) & 0.19 (0.11) & 0.41 (0.2) & 0.32 \\ 
 &  & \hspace{4 mm}+Contr. & 0.34 (0.14) & 0.23 (0.19) & 0.52 (0.24) & 0.37 (0.2) & 0.4 (0.24) & 0.44 (0.18) & 0.38 \\ 
 \cmidrule{2-10}
& \multirow{3}{*}{2.5k} & PEBBLE & 0.15 (0.086) & 0.13 (0.076) & 0.16 (0.1) & 0.16 (0.09) & 0.18 (0.075) & 0.25 (0.11) & 0.17 \\ 
 &  & \hspace{4 mm}+SAF & 0.33 (0.19) & 0.12 (0.082) & 0.32 (0.17) & 0.18 (0.09) & 0.27 (0.11) & 0.22 (0.14) & 0.25 \\ 
 &  & \hspace{4 mm}+Contr. & 0.21 (0.13) & 0.19 (0.22) & 0.29 (0.17) & 0.17 (0.09) & 0.25 (0.15) & 0.28 (0.16) & 0.23 \\ 
\bottomrule
\end{tabular}
\end{sc}
\end{scriptsize}
\end{center}
\end{table*}

We present results ablating the impact of our modified SAF reward network architecture in Table \ref{app_tab:completeperf_ratio_pebble_pebblesfc}, see Section \ref{sec:encoding_env_dynamics}, State-Action Fusion Reward Network for details. In our ablation, we replace the original PEBBLE reward network architecture from \citep{lee2021pebble} and replace it with our SAF network and then evaluate on the joint experimental condition with no other changes to reward function learning. We evaluate the impact of the SAR reward network on the walker-walk, quadruped-walk, sweep into, and button press tasks. Policy and reward function learning is evaluated across feedback amounts and labelling styles. All hyper-parameters match those used in all other experiments in the paper (see Appendix \ref{app_sec:hyper_parameter_details}). We compare PEBBLE with the SAF reward network architecture (PEBBLE + SAF) against SAC trained on the ground truth reward, PEBBLE with the original architecture (PEBBLE), PEBBLE with SimSiam SPR (PEBBLE+SimSiam), and PEBBLE with Contrastive SPR (PEBBLE+Contrastive).

The inclusion of the SAF reward network architecture does not meaningfully impact policy performance. In general, across domains and experimental conditions, PEBBLE + SAF performs on par with or slightly worse than PEBBLE. The lack of performance improvements suggest that the performance improvements observed when the auxiliary temporal coherency objective are due to the auxiliary objective and not the change in network architecture.

\clearpage

\end{document}